# Automatic error detection in part of speech tagging

David ELWORTHY
Sharp Laboratories of Europe Ltd.
Edmund Halley Road
Oxford Science Park
Oxford OX4 4GA
UK
dahe@sharp.co.uk

## Abstract

A technique for detecting errors made by Hidden Markov Model taggers is described, based on comparing observable values of the tagging process with a threshold. The resulting approach allows the accuracy of the tagger to be improved by accepting a lower efficiency, defined as the proportion of words which are tagged. Empirical observations are presented which demonstrate the validity of the technique and suggest how to choose an appropriate threshold.

*Citation details: appears in Proceedings of the International Conference on New Methods in Language Processing, Manchester, September 14-16 1994, pp. 130–135.*

## Introduction

Tagging by Hidden Markov Model (HMM) is a successful technique for assigning grammatical information to words in a corpus. In outline, the tagger picks the most probable grammatical category, or *tag*, for each word by combining the probability of possible tag sequences with the probability of each hypothesised tag for the word, considered in isolation from the context. There are two standard algorithms, known as Forward-Backwards (FB) and Viterbi. Viterbi achieves greater computational efficiency than FB by pruning out some hypotheses; FB has the advantage of assigning a score to every hypothesised tag. The model, in the form of the word-tag (lexical) and tag-tag (transition) probabilities may be estimated from a training corpus which has been tagged by some other means, for example by a human annotator. Alternatively, the model may be derived using a procedure called Baum-Welch re-estimation, which iteratively improves the quality of an approximate model based on probabilities derived by the FB algorithm. The tagging algorithms are robust, in the sense that every word will receive a tag, even in ungrammatical text. For a more detailed description including the mathematical foundations of the technique, see Sharman (1990) and Cutting et al. (1992).

Tagging based on HMMs is capable of achieving a good success rate. Typically, about 90% of words which have more than one tag are assigned the correct part of speech (Cutting et al., 1992), equivalent to about 96% correctness over the whole corpus. While 96% sounds like a good success rate, it still represents one error each 25 words on average, and such accuracies are only achieved where the training and test corpora are of similar vocabulary and style. Furthermore, since the tagger always succeeds in assigning a tag to every word, there is no direct way of knowing where the errors are.

This paper reports some experiments on the automatic detection of tagger errors. The basic idea is to use observable values from the tagging process to provide an indication of whether the chosen tag is likely to be correct. A similar approach can be applied to Baum-Welch re-estimation, in order to screen poor data out of the re-estimation process, although we will not discuss this here.

The suggestion that an appropriate threshold might enable errors to be detected is not a new one; some related work is discussed in a later section. The contribution of this paper is firstly to derive the data that suggests an error detection technique can be made to work, and then to predict and verify the effects of various thresholds. As a consequence, it is possible to construct a tagger in which either the accuracy or the efficiency is supplied as a parameter.

## The experiment

The proposed means of detecting and correcting tagger errors is to mark a tag as being unreliable if some observable value associated with it lies on the wrong side of a threshold. The most obvious observables to use are the probability of the tag and its entropy. Probability measures the likelihood compared to other hypotheses for the same word, and so a low probability may indicate that the tagger has not managed to make a clear disambiguation. Entropy (very roughly) gives a measure of the information content of the tag, which is again an indication of distinctness. To determine the suitability of these observables, the tagger was run on various corpora, and their distributions collected. We first introduce the tagger and corpora used in the experiment and then present the data which resulted.

### The tagger and the corpora

The tagger is an implementation of the Viterbi, Forward-Backwards and Baum-Welch algorithms, written in C++. Three test corpora were constructed using parts of the Wall Street Journal corpus from the Penn treebank (Marcus et al., 1993), referred to as WSJ1, WSJ2 and WSJ3 below. WSJ1 was drawn from parts 4000 and 4100 of the WSJ in the treebank, WSJ2 from parts 0000 to 3900 inclusive and WSJ3 from 4200 and 4300. Each corpus was used to train the tagger separately. WSJ1 was then used as test data against the resulting models. The three tests represent the case where there is a very good match between the training and test data, where the match is less good but the training data is sufficiently large to provide reasonable quality, and where there is a poor match. Words which appeared in the test corpus but not in the training corpus were assigned all non-closed class tags.

The accuracy of the tagger on each test is given in table 1, using the results from the FB algorithm. The *"All"* column is the accuracy over the whole corpus, i.e. the proportion of words for which the tag chosen by the tagger was the correct one, and *"Ambig"* is the corresponding figure for words which have more than one hypothesised tag. The proportion of such word tokens is shown as *"Ambiguity"*. For most of the rest of this paper, accuracies are those of ambiguous words only. This is generally a more interesting figure than the overall accuracy, since it includes only those words where the tagger has to select between hypotheses. In passing, it is interesting to note that the tagger can achieve good performance even on the WSJ2 test, where 3 words out of 4 are ambiguous. Corpora WSJ1 and WSJ3 consisted of about 100,000 words, and WSJ2 of about 2,000,000 words.

### Empirical study

There are a number of observables which might prove to be useful in predicting when the tagger has made an incorrect choice. As we have said, the most obvious is the probability $p$ assigned to a tag by the FB algorithm. A second likely measure is the hypothesis entropy, defined as $-plog_2(p)$, which gives an indication of how much information had to be contributed to make the choice compared with a purely random one. One further parameter to the study is whether to collect data for all hypotheses, or for just the one chosen by the tagger. The former is more useful for attempting to improve Baum-Welch re-estimation, and the latter for tagging itself. The results here are for the chosen tag only.

The results of the tests appear in figures 1 and 2 which show the cumulative frequency distributions of hypotheses with the given measure, for ambiguous tokens. On the left hand graph, the curves from left to right are for correct hypotheses in WSJ2, WSJ1 and WSJ3 and incorrect hypotheses in WSJ1, WSJ2 and WSJ3, respectively; on the right hand one, they are incorrect hypotheses from WSJ3, WSJ2 and WSJ1, and correct hypotheses from WSJ3, WSJ1 and WSJ2, respectively. The scores assigned by the FB algorithm to the hypotheses on a word do not total to 1, and in general are not comparable from one word to another. They were therefore normalised first.

The important point is that the shapes of the graphs are broadly comparable across the three corpora. The WSJ3 test shows the greatest deviation from the others, since the test and training data correspond less well than in the other two tests. The distributions are generally quite "well-behaved"; for example, they are monotonic and more or less smooth. These two facts give some reasonable confidence that a technique which is reliable, parameterisable and applicable across corpora can be developed.

### Thresholding for error detection

The data of the previous section can be used to decide on suitable thresholds on the observable values, such that hypotheses which lie on the wrong side of the threshold are rejected. The ideal threshold is one which eliminates a large proportion of the incorrect hypotheses while retaining as many of the correct ones as possible. In the case of the probability, the threshold is used as a lower bound, i.e. hypotheses with lower probabilities are rejected. For entropy, it forms an upper bound, the inversion arising from the fact that the measured value is negated.

To evaluate the technique, we need a measure of accuracy. Two definitions are possible. If words with a rejected hypothesis are simply ignored, the accuracy is calculated by dividing the number of correctly tagged, non-rejected words by the total number of non-rejected words. Alternatively, if the existence of an "oracle" may be assumed, which will find the correct tag for

*Elworthy*

| Training corpus | All (%) | Ambig (%) | Ambiguity (%) |
|---|---|---|---|
| WSJ1 | 96.93 | 92.66 | 41.85 |
| WSJ2 | 94.20 | 92.85 | 77.87 |
| WSJ3 | 90.18 | 83.07 | 49.97 |

Table 1: Basic tagger accuracy on test corpus WSJ1

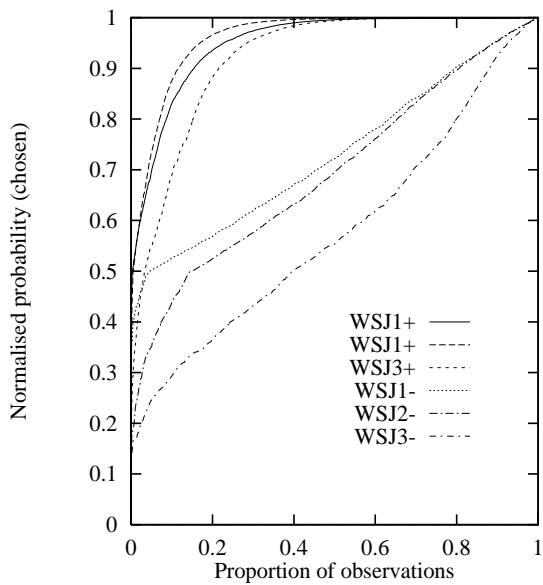

Figure 1: Probability vs. Number of observations

any rejected words, the accuracy is found by summing the proportion of words which are correctly tagged and not rejected, and the proportion of words which are rejected and so handled by the oracle. Suppose $s$ is the proportion of ambiguous words that are correctly tagged, $c$ the proportion of correctly tagged ambiguous words which are (incorrectly) rejected, and $i$ the proportion of incorrectly tagged ambiguous words which are (correctly) rejected. Then the accuracy on non-rejected ambiguous words, i.e. the first method of evaluation, is $s(1-c)/(1-sc-(1-s)i)$, and $s+(1-s)i$ by the second method[1]. A second figure for evaluating the process is the *efficiency* of the tagger, defined as the proportion of ambiguous words actually labelled, whether correct or not. This gives a measure of how hard the oracle must work in the second case. The efficiency is $s(1-c)+(1-s)(1-i)$.

---

[1] This can be seen as follows: $c$ of the words in total are correct. Of these $sc$ are rejected and $s(1-c)$ retained. $1-s$ are incorrect with $(1-s)i$ rejected and $(1-s)(1-i)$ retained. For the "ignore" measure the retained correct words are $s(1-c)$, and the retained words in total $s(1-c)+(1-s)(1-i)$. For the "oracle" measure, $s(1-c)$ correct words are retained and $sc+(1-s)i$ are supplied by the oracle.

As a test to decide which of the proposed measures (probability, entropy) is best, the threshold which yields a specified accuracy was determined using figures 1 and 2, and the corresponding efficiency found. The results are shown in tables 2 and 3, for 95% and 99% (oracle) accuracy on ambiguous words, $P$ standing for probability and $E$ for entropy. With the proportions of ambiguous words as listed in table 1, the success rates across the whole of each corpus (WSJ1, 2 and 3) would be 97.9%, 96.1%, and 97.5% (at 95%) and 99.6%, 99.2%, and 99.5% (at 99%). The tables show that the two measures give similar efficiencies, and that a lower efficiency results with low quality training data (WSJ3). The WSJ1 and WSJ2 tests reject about one word in 20 to obtain 95% accuracy, or one word in 4 (WSJ1) or 5 (WSJ2) to obtain 99%.

Finally, the predictions were confirmed by running the tagger with the thresholds for 95% efficiency. The results are presented in table 4. The experimental results show close correspondence with the predicted ones. This confirms (for the test data, at least) that if we have evidence that allows a threshold to be chosen, then its effect can be predicted, allowing a controlled tradeoff between efficiency and accuracy.

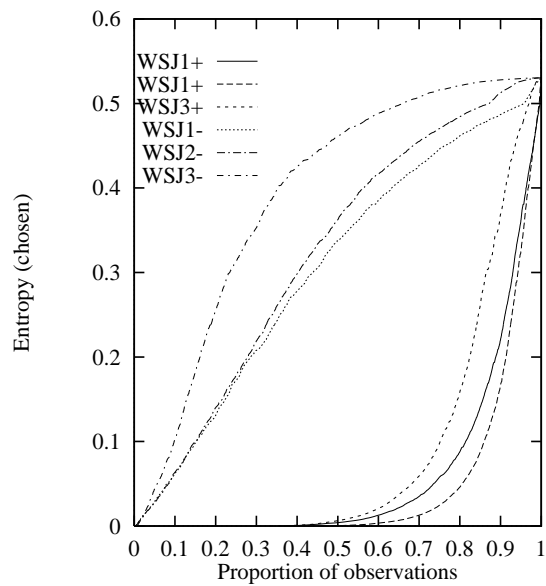

Figure 2: Entropy vs. Number of observations

| Measure | Training corpus | 95% accuracy | |
|---|---|---|---|
| | | threshold | efficiency (%) |
| P | WSJ1 | 0.629 | 94.8 |
| E | WSJ1 | 0.420 | 94.9 |
| P | WSJ2 | 0.577 | 96.2 |
| E | WSJ2 | 0.455 | 96.2 |
| P | WSJ3 | 0.710 | 79.2 |
| E | WSJ3 | 0.349 | 79.2 |

Table 2: Efficiency for 95% accuracy on ambiguous words

A follow-up experiment was conducted using the same thresholds and training corpora, but with WSJ3 as the test corpus rather than WSJ1. When training with WSJ2 the results were broadly comparable to the figures above, and with WSJ1 the results are similar to the values obtained with WSJ3 in the previous test, as would be expected. With WSJ3 as training data, which could be expected to compare with the case of WSJ1 before, shows a poorer correspondence, suggesting that when we move to a significantly different corpus, then the frequency distributions must be collected afresh if the accuracy and efficiency are to be predicted accurately.

## Related work

There are a number of references in the literature to the use of thresholds. Weischedel et al. (1993 p.367) uses thresholds to eliminate tags from unknown words when their probability is less than $1/e^2$ that of the most likely tag. Garside et al. (1987 p.51) apply thresholding to the output of the tagger CLAWS1, and retain tag ambiguities when the tag probabilities are judged too low to indicate a reliable result. Neither of these sources include information about the distribution of probabilities and the accuracy-efficiency tradeoff.

A examination of automatic error detection in more detail can be found in Macklovitch (1992) and the related work in Foster (1991). Macklovitch reports a comparison between two experiments on error detection and correction, one based on statistical information and the other linguistic knowledge. The statistical approach used an error detection technique developed by Foster, which assumes that "tagging errors correspond to situations in which the model assigns similar probabilities to competing alternatives". He therefore compares the differences between scores on the best and second best hypotheses with a threshold. This can be seen to be roughly equivalent to the approach described above, as follows. Let $p$ be the probability of the chosen tag. In the case where there are two hypothesised tags, the other tag has probability $1 - p$. Then the difference in the probabilities is $p - (1 - p) = 2p - 1$, and hence comparing this against a threshold $n$ is equivalent to comparing $p$ against $(n+1)/2$. Similar remarks apply if the ratio is used rather than the difference. If there are more than two hypothesised tags, the equivalence between the two techniques is less exact, but in two common cases – where all hypotheses have similar probability, and where the third and remaining hypotheses all have very low probability – similar reasoning will apply. Macklovitch reports the statistical technique as having similar performance to the results quoted above; for example, it will detect all but 1% of the tagging errors while rejecting 15% of the tokens, i.e. 99% efficiency and 85% accuracy.

| Measure | Training corpus | 99% accuracy | |
|---|---|---|---|
| | | threshold | efficiency (%) |
| P | WSJ1 | 0.935 | 75.2 |
| E | WSJ1 | 0.888 | 75.2 |
| P | WSJ2 | 0.932 | 80.3 |
| E | WSJ2 | 0.0936 | 80.4 |
| P | WSJ3 | 0.963 | 57.6 |
| E | WSJ3 | 0.516 | 57.9 |

Table 3: Efficiency for 99% accuracy on ambiguous words

| Measure | Training corpus | Threshold | Accuracy (%) | Efficiency (%) |
|---|---|---|---|---|
| P | WSJ1 | 0.629 | 94.97 | 94.9 |
| E | WSJ1 | 0.420 | 95.01 | 94.8 |
| P | WSJ2 | 0.577 | 94.99 | 96.2 |
| E | WSJ2 | 0.455 | 95.01 | 96.1 |
| P | WSJ3 | 0.710 | 94.98 | 79.2 |
| E | WSJ3 | 0.349 | 95.04 | 79.1 |

Table 4: Measured accuracy and efficiency, WSJ1

Macklovitch's linguistic, or knowledge based, technique uses rules to flag potentially incorrect tagging of past and preterite verb forms, a common tagger error. The rules are based on linguistic principles; for example, a word tagged as participle is marked as being an error if it is the only verbal form in a sentence. 68% of the tags flagged by the rules were genuine errors, the remainder being spuriously flagged. Of the past/preterite errors made by the tagger, about 30% were detected by the rules. The statistical approach detected 49% of the tagger's errors when adjusted to give the same rate of errors in the detection.

The major argument in favour of using linguistic rules is that it is possible to specify how to correct the error as well as to detect it. However, there are some disadvantages to the knowledge-based approach. It is not parameterisable, in that one cannot choose a threshold to give a specified efficiency or accuracy. A second problem is that it introduces knowledge about the constructs of a specific language into the tagger, which is otherwise (barring the design of the tagset) language independent. The knowledge is only applicable on the assumption of reasonably grammatical input. However, one of the strengths of HMM tagging is that such requirement can largely be avoided in favour of an assumption that the training and test data are similar. Finally, it is likely that if a wide range of errors are to be detected, then many rules covering the different sources of error would be needed. The interactions between these rules might prove difficult to control.

## Applications

Although we have yet to apply the techniques described here in an anything other than an experimental system, a possible application area is as follows. In the Integrated Language Database project[2] we are developing systems which will allow lexicographers and computational linguists to construct lexicons in which the lexical data is supported by evidence from analysed corpora. It is certainly reasonable to expect the corpora to be tens or hundreds of millions of words, providing instances of most lexical items in varying contexts. When using such data, it is more important to the lexicographer or computational linguist to know that the citations on which they are basing their decisions are reliable, even if this results in only a few examples. Consequently, the tagger can be run with a threshold which provides

---

[2]The ILD project is a collaboration between Sharp Laboratories, Cambridge University Press, Cambridge University Computer Laboratory and Edinburgh Language Technology Group, and part funded under the UK Department of Trade and Industry's SALT programme.

a certain degree of accuracy, accepting a reduced efficiency as a consequence.

## Conclusions

We have shown the results of applying thresholds to hypothesis scores during tagging. Thresholding allows the accuracy of the tagger to be increased, at the cost of a reduction in efficiency. Depending on the quality of the training data, accuracies of 95% on ambiguous words can be achieved at efficiencies of 80%–95%, or accuracies of 99% at efficiencies of 57%–80%.

Two suggestions for extending this work are to collect data on probabilities for specific tags, as Garside et al. (1987) do for CLAWS1, and to examine the probability distribution on models (i.e. sequences of ambiguous words) containing specific numbers of errors, with the view to selecting different thresholds in different circumstances. These techniques potentially allow better error detection, but at the cost of losing the simple relation between accuracy and efficiency and the corresponding tunability of the tagger.

## Acknowledgments

An earlier version of the techniques was tried out at Cambridge University Computer Laboratory, as part of Esprit BR Project 7315 "The Acquisition of Lexical Knowledge" (Acquilex-II). Ted Briscoe made a number of useful comments on those experiments. The work reported in the paper was carried out at Sharp Laboratories of Europe. I also thank the NeMLaP reviewers for their comments.